\documentclass{article}

\usepackage{PRIMEarxiv}

\usepackage[utf8]{inputenc} 
\usepackage[T1]{fontenc}    
\usepackage{hyperref}       
\usepackage{url}            
\usepackage{color}
\usepackage{booktabs}       
\usepackage{amsfonts}       
\usepackage{nicefrac}       
\usepackage{microtype}      
\usepackage{lipsum}
\usepackage{fancyhdr}       
\usepackage{fancyvrb}
\usepackage{graphicx}       
\usepackage{amsmath}
\usepackage{mhchem}
\usepackage{makecell}
\usepackage[ruled,vlined,linesnumbered]{algorithm2e}
\usepackage{upgreek}
\usepackage[backend=biber,style=numeric,maxnames=3,minnames=1,sorting=none]{biblatex}
\usepackage{textcomp}

\newcommand{\micron}{$\upmu$m}

\title{EAA: Automating Materials Characterization with Vision Language Model Agents}

\author{
  Ming Du$^*$ \\
  Advanced Photon Source \\
  Argonne National Laboratory \\
  Lemont, IL, USA \\
  $^*$ \texttt{mingdu@anl.gov} \\
\AND
  Yanqi Luo \\
  Advanced Photon Source \\
  Argonne National Laboratory \\
  Lemont, IL, USA \\
\AND
  Srutarshi Banerjee \\
  Data Science and Learning Division \\
  Argonne National Laboratory \\
  Lemont, IL, USA \\
\AND
  Michael Wojcik \\
  Advanced Photon Source \\
  Argonne National Laboratory \\
  Lemont, IL, USA \\
\AND
  Jelena Popovic \\
  Department of Radiation Oncology \\
  Northwestern University \\
  Chicago, IL, USA \\
\AND
  Mathew J. Cherukara$^\dagger$ \\
  Advanced Photon Source \\
  Argonne National Laboratory \\
  Lemont, IL, USA \\
  $^\dagger$ \texttt{mcherukara@anl.gov}
}

\begin{document}

\textbf{GOVERNMENT LICENSE}

The submitted manuscript has been created by UChicago Argonne, LLC, Operator of Argonne
National Laboratory (``Argonne''). Argonne, a U.S. Department of Energy Office of Science laboratory, is operated under Contract No. DE-AC02-06CH11357. The U.S. Government retains for
itself, and others acting on its behalf, a paid-up nonexclusive, irrevocable worldwide license in
said article to reproduce, prepare derivative works, distribute copies to the public, and perform
publicly and display publicly, by or on behalf of the Government. The Department of Energy will
provide public access to these results of federally sponsored research in accordance with the DOE
Public Access Plan. http://energy.gov/downloads/doe-public-access-plan.

\maketitle

\begin{abstract}
We present Experiment Automation Agents (EAA), a vision-language-model–driven agentic system designed to automate complex experimental microscopy workflows. EAA integrates multimodal reasoning, tool-augmented action, and optional long-term memory to support both autonomous procedures and interactive user-guided measurements. Built on a flexible task-manager architecture, the system enables workflows ranging from fully agent-driven automation to logic-defined routines that embed localized LLM queries. EAA further provides a modern tool ecosystem with two-way compatibility for Model Context Protocol (MCP), allowing instrument-control tools to be consumed or served across applications. We demonstrate EAA at an imaging beamline at the Advanced Photon Source, including automated zone plate focusing, natural language-described feature search, and interactive data acquisition. These results illustrate how vision-capable agents can enhance beamline efficiency, reduce operational burden, and lower the expertise barrier for users.
\end{abstract}


\section{Introduction}

The term ``agent'' is defined as an autonomous system that collects information about the environment and takes actions according to the information collected to realize its objective \cite{Franklin1996-agent}. When allowed to use certain tools such as a search engine, a data query interface or a function that moves an electric motor in its action phase, an agent is more specifically referred to as a ``tool agent'' \cite{Wang2024-na}, with its capability significantly extended. As large language models (LLMs) and particularly vision language models (VLMs) evolve towards more powerful reasoning skills, tool-using capabilities, and image comprehension, a vast potential for applying AI agents in automating interactions with real-world objects is emerging. A report by Deloitte predicts that by 2027, about 50\% of companies using generative AI will have launched pilot projects with autonomous agentic AI \cite{Loucks2024-DeloitteInsights}. The productivity enhancement through the introduction of AI agents to human teams is substantial \cite{Ju2025-kr, Noy2023-gc}. The growing capabilities of LLMs and VLMs are also promoting their adoption in scientific research, with experts foreseeing AI agents accelerating discoveries in biology, medicine, chemistry, and materials science  \cite{Devereson2025-McKinsey}.

At synchrotron light sources, AI agents are becoming increasingly important. Running synchrotron beamlines and carrying out X-ray experiments during beamtime often requires repetitive, time-intensive work such as tuning optics and searching for features of interest. Many of these steps are hard to capture with fixed, rule-based automation because they depend on contextual and semantic interpretation of data and images. At the same time, first-time external users can be overwhelmed by dense and complex instrument control systems, which may not even provide a graphical interface. This steep control barrier drives the need for extensive training, reduces beamtime efficiency, and increases the likelihood of human error. Equipping AI agents with tools that can operate beamline hardware -- such as sample-stage motors, optical actuators, shutters, and detectors -- offers two major advantages: it can take over routine, repetitive procedures, allowing beamline staff to focus on higher-value work; and, through a natural-language chat interface, it can act as an accessible layer that helps inexperienced users run experiments more confidently and effectively. This opportunity has been explored by researchers at several synchrotron facilities over the recent years. The CALMS system introduced in \cite{Prince2024-bj} can answer user questions about experiment design across multiple scientific facilities by leveraging conversation memory and document stores. Additionally, through integration with the Materials Project database API and an API for controlling x-ray diffractometer motors based on a material's lattice constants and desired Bragg peak, CALMS has successfully demonstrated the ability to automatically position a diffractometer at the Advanced Photon Source when a user requests measurement of a specific Bragg peak for a given material. VISION described in \cite{Mathur2025-rr} is a modular framework comprising multiple sub-agents and tools, including a coding sub-agent that composes Python code for experiment control and data processing, to provide voice- and text-controlled data acquisition and analysis. 

We aim to further advance the functionality and productization of AI agents for synchrotron beamline operations by filling the gaps that remain between the current state-of-the-art and the actual complexity in beamline experiments. Below, we identify the areas where improvements can be made and explore the opportunities to close them based on modern LLM/VLM technology and the experience accumulated in both industry and academia over the course of AI adoption:

\begin{itemize}
    \item \emph{Image comprehension}. Many beamline operation tasks require vision-based decision-making, and semantic understanding of images is often necessary. For example, locating a small feature with a certain characteristic (\emph{e.g.}, ``a hexagon-shaped nanoparticle'') in a large sample may require one to run repeated local image acquisitions with at least moderate resolution (and thus a small field of view) and examine each image to tell if the desired feature is present. The critical step in this task is to detect the feature in an image given the description of that feature in natural language. Semantically associating language and image has been a major challenge for agents based on text-only LLMs, but the development of vision language models (VLMs) supporting multi-modal inputs, examples of which include Claude Opus, ChatGPT-5 and Gemini 3, has provided an effective solution to this problem. We recognize that the ability to comprehend images from both users and tools in line with text prompts is crucial for AI agents deployed at synchrotron beamlines. 

    \item \emph{Friendliness with logic-based workflows and analytical routines}. One of the key advantages of LLM agents is their ability to make flexible, adaptive decisions without relying on rigid rules. However, hallucination remains a significant challenge—where these models produce inaccurate information or fabricate non-existent concepts—raising serious concerns about both their effectiveness and safety. \cite{Huang2025-rp}. While tools that perform data retrieval or numerical computation can be used by the LLM to make well-supported and quantitatively accurate decisions, uncertainty still exists in whether the LLM calls the right tools and passes in the right arguments, and the tool calling performance of LLMs is commonly seen to degrade when too many tools are given. Moreover, some applications already have highly optimized logic-based workflows or analytical routines, such as Bayesian optimization \cite{Du2025-bs}, that steers the process. Allowing an LLM agent to take over the entire workflow and replacing all analytical routines embedded in the workflow with tool calls are not guaranteed to yield results comparable to traditional approaches, and undermine the predictability of the operations. We believe a production-robust agentic system should offer the possibility to build processes at all levels of balance between LLM control and logic-based routines: for interactive sessions or tasks involving fuzzy decision-makings, the LLM agent can drive the process and use tools at its disposal; for more complicated tasks, certain analytical routines should be triggered by rules instead of tool call; in a more extreme scenario, the high-level workflow should be defined by optimized logic while queries to the LLM are embedded in the process as subroutines. 

    \item \emph{Modernized tool standard}. A proper set of tools is essential for agents to complete their tasks. Function calling is an early and convenient form of tool-using for agents, where the tools are program functions. Tool calls made by the agent are parsed and the called function is executed on the application side. While easy to implement, function calling tools are usually usable only with the agent application where they are defined, because there does not exist a standardized interface for function calling tools to be consumed across different applications. The model context protocol (MCP) \cite{MCP2024} introduced in 2024 defines a universal standard for agent-tool interface. An MCP tool is set up as a standalone program that exposes endpoints for input and output, also known as an MCP server; agent applications that use MCP tools are called MCP clients. As a broadly accepted standard, MCP tools can be consumed by clients developed by different developers or organizations supporting MCP, with the clients being completely agnostic about the internal definition and implementation of the tools. To facilitate the construction of a self-consistent ecosystem, a modern agentic system may have its own library of function calling tools but should also provide two-way support for MCP tools: it should be able to use external MCP tools, and should also be able to serve its own tools as MCP servers for other clients to consume. 

    \item \emph{Long-term memory}. Conversational memory of an agent relies on the context, which is ephemeral and vanishes after the session ends. Long-term memory, on the other hand, allows the agent to build and accumulate a knowledge base through its interaction with the user. This is commonly realized through retrieval-augmented generation (RAG).

    \item \emph{Compatibility with instrument control library.} Some agentic AI frameworks execute tools concurrently, for example by running multiple tool calls in parallel threads or asynchronous tasks. This can create reliability issues when tools interface with scientific instruments through control libraries that are not designed to handle overlapping requests, often because they maintain internal state, rely on native drivers, or assume a strictly sequential command flow. Running tools as external services (for example, as MCP servers) in dedicated processes provides an effective way to isolate internal state and avoid thread-safety problems, and is therefore an important capability of our system. However, such process-based isolation requires either maintaining long-running server processes or launching tool processes on demand; the latter limits the ability of tools to preserve state across calls. For this reason, it is desirable to also support tools that are invoked through function calling within the agentic runtime, where tool instances can persist in memory and maintain state over time. Supporting both execution models means that thread safety cannot be addressed solely through process isolation and must also be considered in the design of the agentic application itself. In addition, true parallel execution of tool calls is often unnecessary or even unsafe for physical instruments, which typically operate sequentially; uncoordinated concurrent actions can lead to conflicts or unintended behavior. As a result, prioritizing controlled, deterministic tool execution over maximal concurrency generally leads to more robust and reliable experimental automation.
\end{itemize}

Aiming for a cutting-edge experiment automation agentic system possessing the above characteristics, we developed EAA, or Experiment Automation Agents. EAA works with VLMs and allows images to be sent to the VLM from both users and tools (tools returning images are not directly supported by most agentic frameworks available in the community). In addition to interactive communications with the user, EAA also allows one to build automation workflows either with the LLM driving the process or with a part or the entire workflow controlled by code logic. Additionally, EAA maintains a library of tools for instrument control, which can both be used by EAA itself as function calling tools and be wrapped into MCP servers so that they can be used by other MCP clients such as Claude Code and Gemini CLI. At the same time, EAA can use MCP tools from other developers. Moreover, EAA can have long-term memory by building and accumulating a vector database through its interaction with the user and searching the database for RAG. Additionally, by avoiding unnecessary multi-threading in tool execution, thread safety is reinforced when EAA works with in-process tools built on instrument control libraries.

In the rest of this paper, we introduce the design of EAA in a top-down approach, beginning with an overview of its high-level architecture and follow it with details about every component. We then present several experimental demonstrations at a complex synchrotron beamline showing the user interaction and automated task execution of EAA.

\section{Software design}

\begin{figure}
    \centering
    \includegraphics[width=0.9\linewidth]{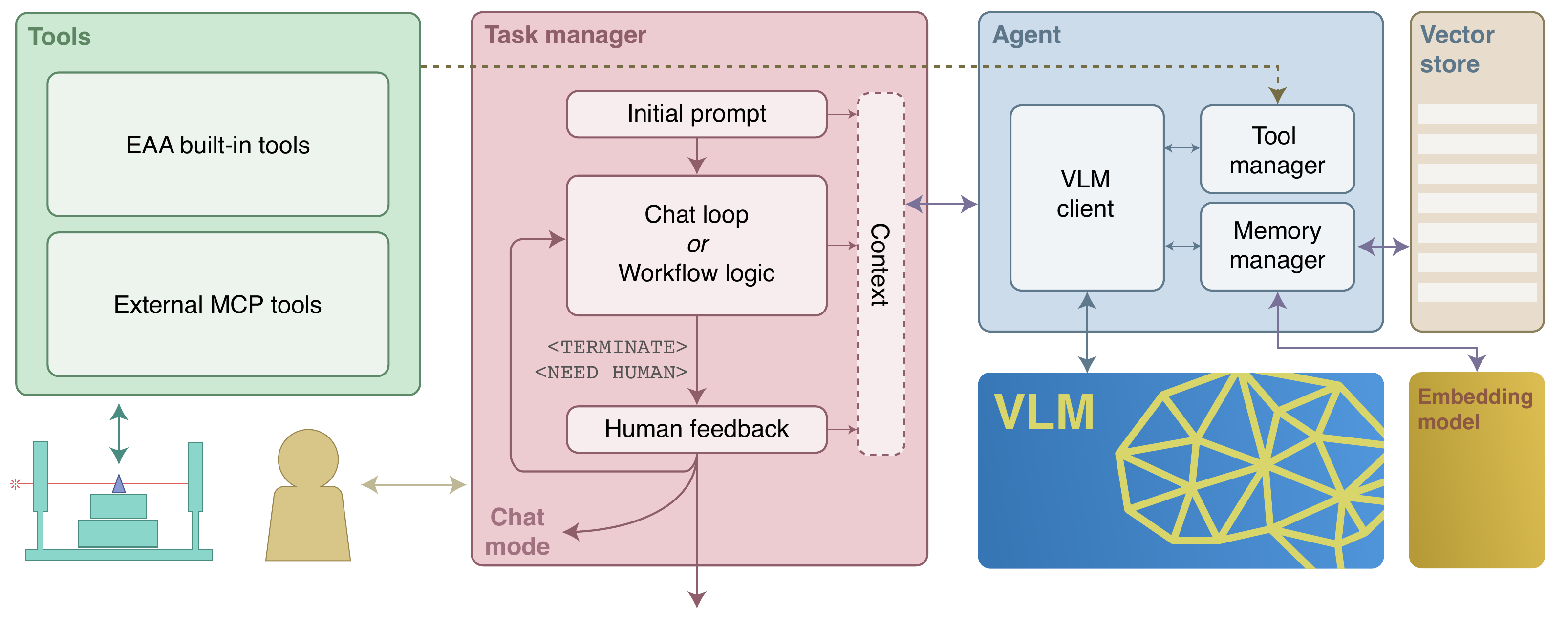}
    \caption{The main components of EAA and their interactions. The task manager contains the chat loop or workflow, creates and holds the agent object, and maintains the context. New messages coming from the user, auto-generated by the workflow logic, or responded by the agent are added to the context. When the LLM agent drives the workflow, it sends a termination signal when the task is considered finished; The task manager then requests human input, based on which it either continues the workflow with updated instructions, enters the purely interactive chat mode, or exits. When a message is sent to the agent, it is sent with the entire context so that the VLM has conversational memory. Optionally, the agent can have long-term memory by saving the embeddings of notable messages into a vector store and retrieving them for future conversations and sessions. Tools from EAA's tool library or provided through external MCP servers are registered to the agent's tool manager through the task manager. The agent generates the schemas for the tools and attach them with messages sent to the VLM, and handles the tool calls made by the VLM.}
    \label{fig:architecture}
\end{figure}

The software architecture of EAA comprises three main modules each with their own class hierarchy: the task manager, the agent, and the tool library. When performing a task, a task manager, an agent and a selection of tools are instantiated. Fig.~\ref{fig:architecture} illustrates their interactions. 

\subsection{Task manager}
\label{sec:task_manager}

\begin{figure}
    \centering
    \includegraphics[width=0.75\linewidth]{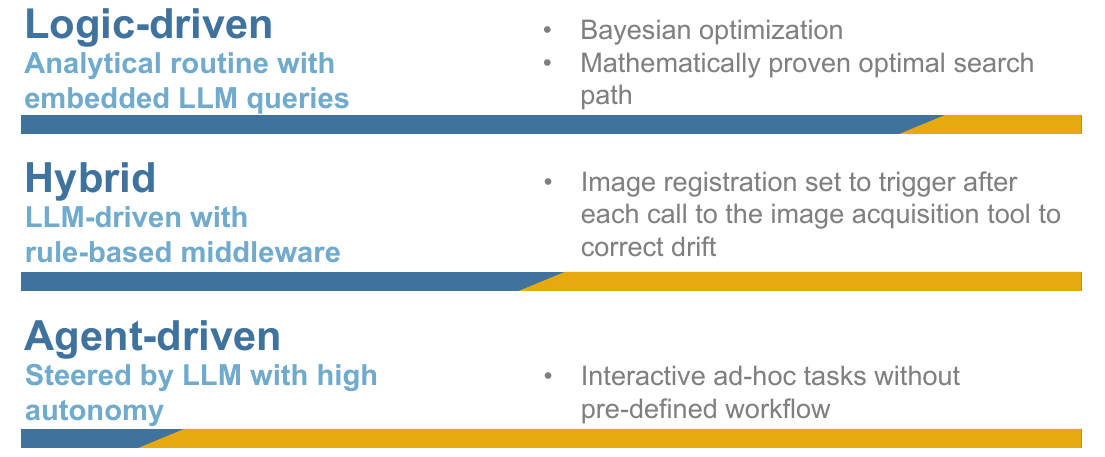}
    \caption{Three levels of LLM involvement in experiment automation tools. Examples are enumerated for each level.}
    \label{fig:three_levels}
\end{figure}

The task manager defines the chat loop realizing the conversational interaction between the user and the agent. Alternatively, it can also implement workflows driven by either the agent or logic. We enumerate three types of workflows with different levels of LLM involvement that are equally well supported by EAA (also illustrated in Fig.~\ref{fig:three_levels}):
\begin{enumerate}
    \item \emph{A logic-driven workflow}. It uses arbitrary Python code rather than the feedback loop to drive the process. Queries to the LLM are embedded in the workflow as localized subroutines for certain steps of the workflow (for example, to determine whether two images with different modalities contain the same feature), with prompts instructing the LLM to respond in a structured format (``yes'' or ``no'', or a JSON string). The response is parsed to guide the next step of the workflow. Important logging information and routine outcomes can still be added into the context of the LLM, so that when the workflow pauses or finishes, the LLM can step in to help analyze or troubleshoot the process. 
    \item \emph{A hybrid workflow}. While this workflow also runs the loop, analytical routines can be inserted at certain points of it or be triggered by defined criteria. The workflow can automatically generate user-proxy messages with the outcomes of the analytical routines and send them to the LLM. To realize this, the feedback loop routine of EAA accepts ``hook functions'' and execute them when conditions are met. 
    \item \emph{An agent-driven workflow}. Starting from an initial prompt from the user which can be generated from a template, the workflow runs in a loop where it sends the prompt to the LLM and receives the response; If the response contains tool calls, the tools are run, and the results are returned to the LLM. When no tool calls exist in the LLM response, the conversation is handed back to the user. The loop implemented in EAA can optionally force the LLM to either make tool calls or send an explicit termination signal (like ``\texttt{TERMINATE}'') in its response; otherwise, the routine sends an auto-generated message to the LLM reminding it to make tool calls, thereby keeping the loop running and preventing premature termination. When a new message is created by any role, it is appended to the context list maintained by the task manager. A message sent to the agent is always sent together with the context in order for the VLM to have conversational memory.
\end{enumerate}

Task managers can be nested to enable sub-agents. A task manager can be created in another task manager with its own agent object (with a different system prompt or model) and context. This allows one to assign sub-agents to handle specific components of a complex task and avoids always passing the entire, fast-growing context to one agent. 

\subsection{Agent}
The agent is responsible for multimodal reasoning, decision making, and tool orchestration. Conceptually, it mediates between the VLM inference endpoint and the experimental capabilities exposed through tools. For each interaction, the agent receives the full conversational context, selects appropriate actions (tool calls or user-facing responses), and returns structured outputs that drive either continued automation or human interaction. Optional long-term memory enables the agent to accumulate persistent knowledge about beamlines and procedures across sessions.

\subsection{Tools}
Tools provide the agent with controlled access to experimental capabilities such as image acquisition, motion control, and optics adjustment. From a high-level perspective, tools define the action space of EAA and encapsulate instrument operations behind well-defined interfaces. EAA supports both in-process tools and externally hosted tools via the Model Context Protocol (MCP), enabling interoperability across applications while preserving robustness for instrument control.

The tool library of EAA contains tools for instrument interactions including performing image scans, moving sample stages, and adjusting optics. The tools of EAA are stateful, where states or data (such as an acquired image) remain persistent across tool calls. EAA's MCP wrapper can convert any of its tool classes into an MCP server so that these tools can be used by other MCP clients. Meanwhile, EAA can also use external MCP servers. This two-way MCP compatibility allows EAA to be used along with software packages in a large ecosystem. 

\subsection{Chat loop and workflows}
\label{sec:workflow}

\begin{figure}
    \centering
    \includegraphics[width=1\linewidth]{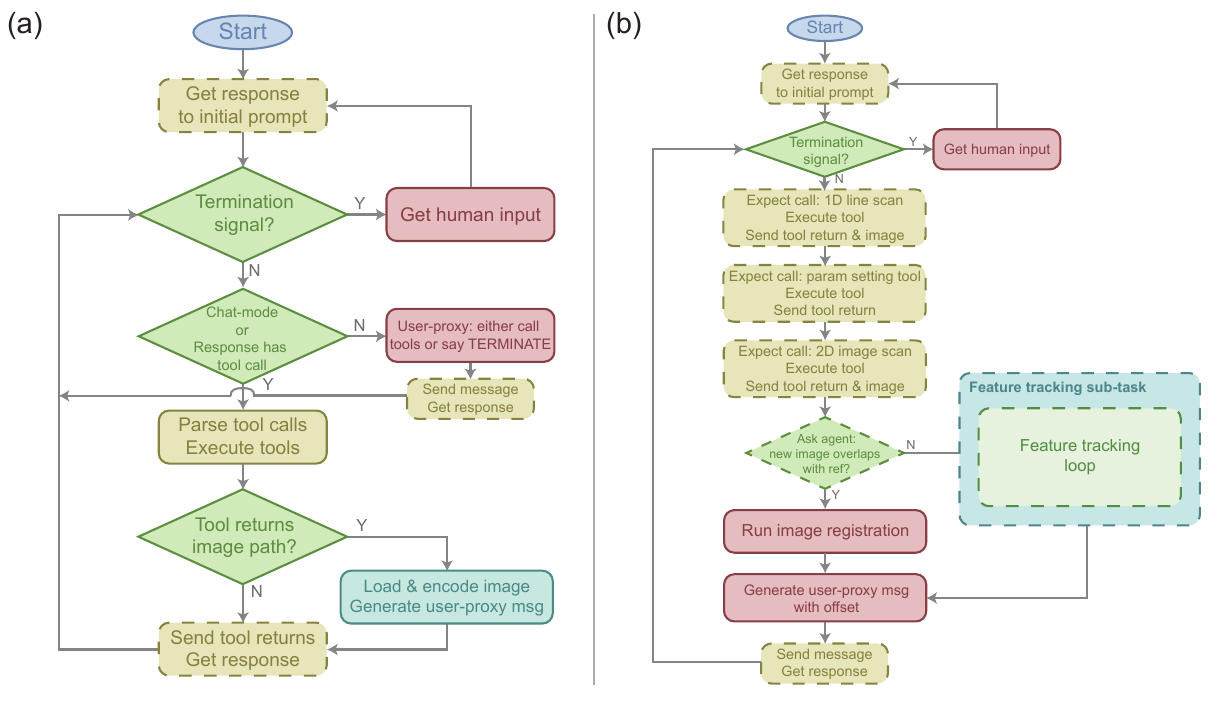}
    \caption{Example workflow diagrams of EAA task managers. (a) The chat loop, which is a generic workflow used for interactive conversations and agent-driven workflows. (b) The workflow of an automated focusing task as an example. This workflow falls in the hybrid workflow category defined in Section \ref{sec:task_manager}, and involves programmed query with the agent about image overlap, logic-based registration and auto-reply composition, and call to a sub-task for feature tracking. In both diagrams, nodes with dashed borders are those involving agent interactions.}
    \label{fig:workflow_diagrams}
\end{figure}

Interactive conversations and agent-driven workflows are handled by the feedback loop in the basic task manager. We use a generic workflow presented in Fig.~\ref{fig:workflow_diagrams} for these scenarios. The agent is queried with an initial prompt that contains a question or instructions for a certain task. When a multi-step task rather than a turn-by-turn conversation is desired, the agent is instructed to explicitly request human input by adding keywords such as \texttt{TERMINATE} or \texttt{NEED HUMAN} in its response; otherwise, a tool call will be expected, and if no tool call is made, an auto-reply will be generated to insist that the agent either sends a termination signal or makes tool calls. This mechanism effectively keeps the agent running as persistently as possible, preventing premature termination. If the response contains tool calls, the tool calls are parsed, and the tools are executed. The tool returns are converted to a tool message in OpenAI-compatible format and added to the context. In particular, EAA supports tools that yield images as outputs, enabling visual feedback from instrument actions to be incorporated into the agent’s reasoning loop. Any additional modality handling required by the underlying API is abstracted away from the workflow logic, allowing the agent to reason over tool outputs in a uniform manner. The tool message, possibly along with an image-containing auto-reply, is sent back to the agent, and the loop continues until a termination signal is captured. 

Some complex tasks are performed using hybrid workflows defined in Section \ref{sec:task_manager}, where analytical routines are inserted into the feedback loop and are triggered by code logic instead of tool calling. We illustrate this with an example of an automated optics focusing task in Fig.~\ref{fig:workflow_diagrams}(b). The agent is briefed with the procedure of optics focusing in the initial prompt given to it. At the beginning of each iteration of the loop, the agent is expected to call the tool that performs a 1D line scan across a designated thin feature (this feature can be presented to the agent in the initial prompt as a reference image or described verbally). The tool will plot the line scan, and fit the intensity profile of the scanned thin feature with a Gaussian peak; the full width at half maximum (FWHM) of the Gaussian will be indicated in the plot, which can be used as a sharpness metric (the smaller the FWHM, the sharper the image). The agent gets the image and notes down the FWHM in its response. In the same response, it is expected to call the tool that sets optics parameters (such as the $z$-position of a zone plate). As the change of the optics will likely cause the image to drift, the agent is asked to acquire a 2D image of the region-of-interest, and check if the image overlaps with the image acquired in the previous iteration. If it does, phase correlation-based image registration is performed by the workflow to obtain the relative offset between the images. Otherwise, a sub-task manager is spawned and asked to perform ``feature tracking,'' where it moves and zooms in/out the field of view to re-align it with the previous 2D image, thereby obtaining the spatial offset. The offset attained with either approach is given back to the agent in the main task manager, which it uses to adjust the location of the next line scan so that the line scan is always performed across the same feature. The process is repeated until a set of parameters that minimize the line scan FWHM is found. The query about image overlap and the execution of image registration or feature tracking are embedded in the workflow through a hook function, which runs automatically when a new 2D image is acquired. 

The above process exemplifies the additional guidance provided by the workflow logic. First, we enforce an expected sequence of tool calls (line scan -- parameter setting -- 2D image acquisition). We use a soft guardrail to balance workflow adherence and flexibility: when the actual order of tool calls does not match the expected sequence, we issue a warning to the agent, but allow it to ignore the warning if it deliberately altered the order to address an exception (\emph{e.g.}, retrying a failed image acquisition). Second, the in-workflow image registration step ensures that offset information is reliably provided when 2D image acquisition is done, making the workflow more predictable than if the agent is expected to call a registration tool. The step of asking the agent to determine whether the current image overlaps with the last one is another example showing how agent interactions can be seamlessly embedded into the workflow, harvesting the vision capability of VLMs to augment the workflow logic. Lastly, we use a sub-agent for the feature tracking task when it is needed. The sub-agent has its own context that is separated from the context used by the main agent, which prevents uncontrolled context growth. 

\subsection{User interface}

EAA has both a command-line interface and a web user interface (WebUI) to display messages from the agent, the workflow (auto-generated), and the human user. In the WebUI, images yielded from tools and sent by users are shown in-line with each message and are also displayed in a dedicated side panel. Users may also take screenshots to their clipboard and paste the image into the input box. This feature is particularly useful for bridging existing beamline software and workflows with EAA. Readers are invited to the videos in the supplementary materials that demonstrate the WebUI in action.

\section{Results}

In this section, we present the intermediate processes and results of process automation and interactive data acquisition using EAA. We used GPT-5 as the VLM. All experiments were performed at beamline 2-ID-D of the Advanced Photon Source using the x-ray fluorescence microscopy (XRF) setup \cite{Davis2022-vb}. We also invite the readers to watch the video recordings of the experiments available in the supplementary materials.

\subsection{Case study 1: microscope focusing}

\begin{figure}
    \centering
    \includegraphics[width=1\linewidth]{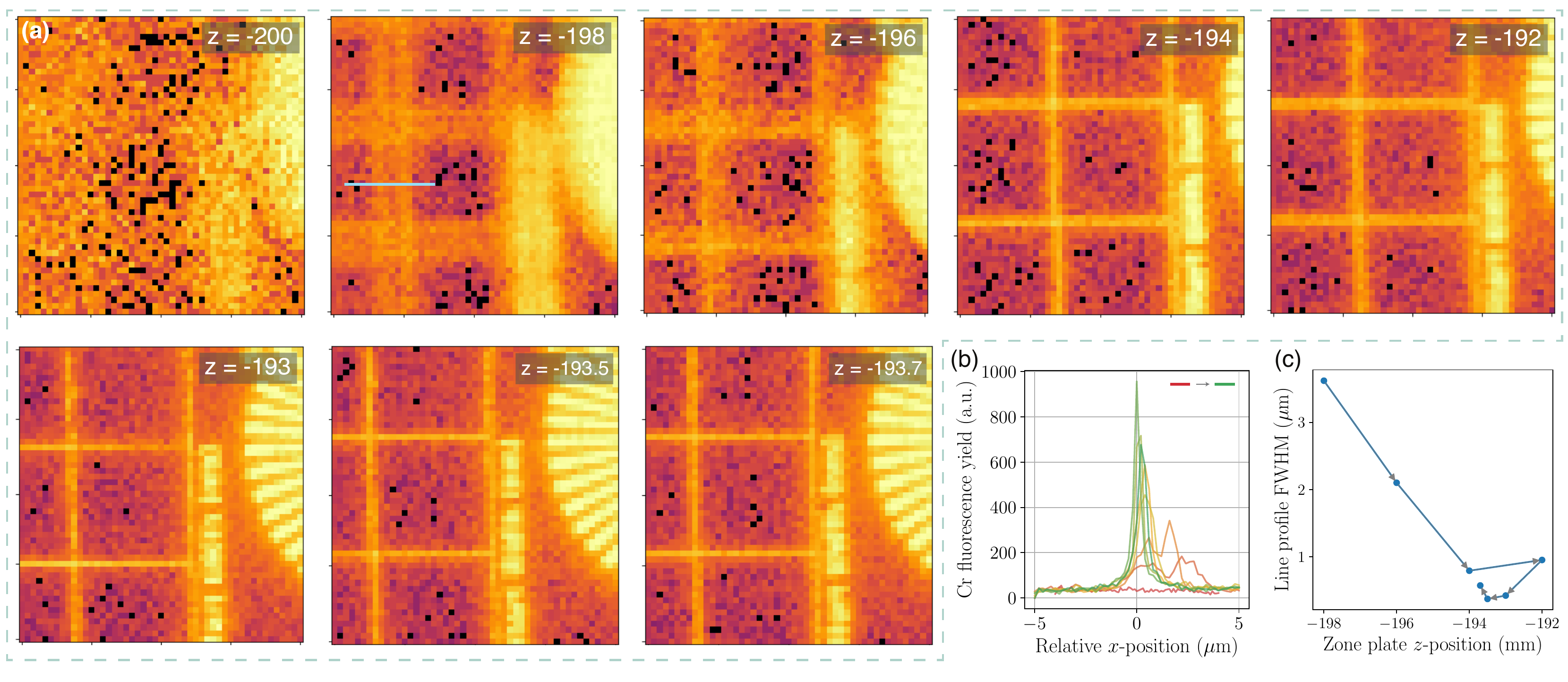}
    \caption{Trajectory of the focusing task. (a) 2D images acquired at all visited zone plate $z$-positions, used for estimating positional drift. The $z$-positions in mm are indicated at the top right corner of each image. The blue line in the image at $z = -198$ indicate the vertical line that the agent was instructed to take line scans across. As the focus of the zone plate improves, images gradually become sharper. (b) The line profiles collected across the reference feature throughout the process. From red to green, the zone plate $z$-position moves from the initial -200 mm towards the final -193.7 mm. The positions of the line plots are relative, meaning the center of the profiles are aligned at $x = 0$ in the figure. (c) The trajectory of the process in the space of line profile FWHM and $z$-position. The initial point at $z = -200$ mm is not shown because its FWHM is too large.}
    \label{fig:focusing_result}
\end{figure}

We first demonstrate an automated zone plate focusing process using the built-in workflow of EAA [Fig.~\ref{fig:workflow_diagrams}(b)]. The task uses a XRF focusing target that has chromium (Cr) features deposited on silicon nitride membrane. The patterns of the test target are composed of a Siemens star and a Cartesian grid of horizontal and vertical lines around it. The lines serve as good reference features, as the dispersion of the peak in the line profiles across them indicates the sharpness of the image. 

Starting from a guessed $z$-position of the zone plate (-200 mm), the agent was instructed to find a $z$-position that forms the sharpest image using the following procedure:
\begin{enumerate}
    \item Acquire a 2D image in the user-specified region. This informs the agent of the structure of the region of operation, and saves the image in the image acquisition tool so that it can be registered with the next image to estimate spatial drift. 
    \item Perform a 1D line scan using the 1D scan tool across the user-described feature. In this case, we specify the feature to be the vertical line at a certain $x$-position, so the line scan should be horizontal. \label{enum:line_scan}
    \item The line scan tool returns a plot of the scanned profile. The tool also automatically fits a Gaussian to the profile and indicates the full width at half maximum (FWHM) of the Gaussian in the figure. Note down the FWHM.
    \item Use the parameter setting tool to change the $z$-position of the zone plate. This would cause the FOV to drift.
    \item Acquire a new 2D image at the same coordinates. The image is sent back to the agent along with a message informing it of the offset between the current image and the last image found by image registration. \label{enum:new_2d_scan}
    \item Go back to step \ref{enum:line_scan}, but perform the new line scan at the coordinates modified by the drift values given in the previous message, so that the relative position of the peak corresponding to the vertical stays unchanged in the profile. Repeat the process until a satisfactory focus is found, which is indicated by a minimal FWHM.
\end{enumerate}

The procedure above mimics how a human beamline scientist would focus a zone plate. The metric of line scan FWHM, although subject to noise and uncertainty, is fast to evaluate. We use the soft guardrail introduced in Section \ref{sec:workflow} to enforce the sequence of the tool calls (line scan $\rightarrow$ parameter setting $\rightarrow$ 2D scan), where a warning message (instead of a hard interruption) is issued to the agent. This preserves the space for the agent to handle unexpected scenarios. As can be seen in the video in the supplementary materials, this warning was triggered during our experiment when the agent re-called the line scan at $z$ = 200 \micron{} because the first line scan's plot did not contain a discernible peak, which is expected due to the bad focus of the zone plate. 

The image registration in step \ref{enum:new_2d_scan} is set to run automatically after the agent collects a new 2D image. The registration is performed using phase correlation and is programmed to run automatically after the image acquisition tool is called. This reduces the workflow's reliance on LLM tool calling, making it more deterministic. 

Including the initial zone plate position at -200 mm, the agent visited 8 positions following the workflow. Fig.~\ref{fig:focusing_result}(a) shows the 2D images acquired at the region of operation at each $z$-position. As the agent moves the zone plate towards better focus, the sharpness of the image gradually improves. Fig.~\ref{fig:focusing_result}(b) shows the line scan profiles obtained at these $z$-positions. The profiles were measured across the vertical line indicated by the blue horizontal line in the image at $z$ = -198 mm in Fig.~\ref{fig:focusing_result}(a). From -200 mm (red) to the final position (green), the line profiles become increasingly sharper. The plot of the experiment trajectory in the space of zone plate $z$-position and the FWHM of the line scan visualizes how the agent approaches the optimal focus (the point at $z$ = -200 mm is not shown due to its overly large FWHM): The agent initially reduces $z$ with a step size of 2 mm. At $z$ = -192 mm, the FWHM was found to increase, so it inferred that the optimum lay between -194 mm and -192 mm and conducted finer search in that interval. While stepping in the negative direction, the FWHM at -193.7 mm was found higher than -193.5 mm, leading the agent to conclude that the optimal focus is at -193.5 mm.

\subsection{Case study 2: searching for desired features}

\begin{figure}
    \centering
    \includegraphics[width=0.75\linewidth]{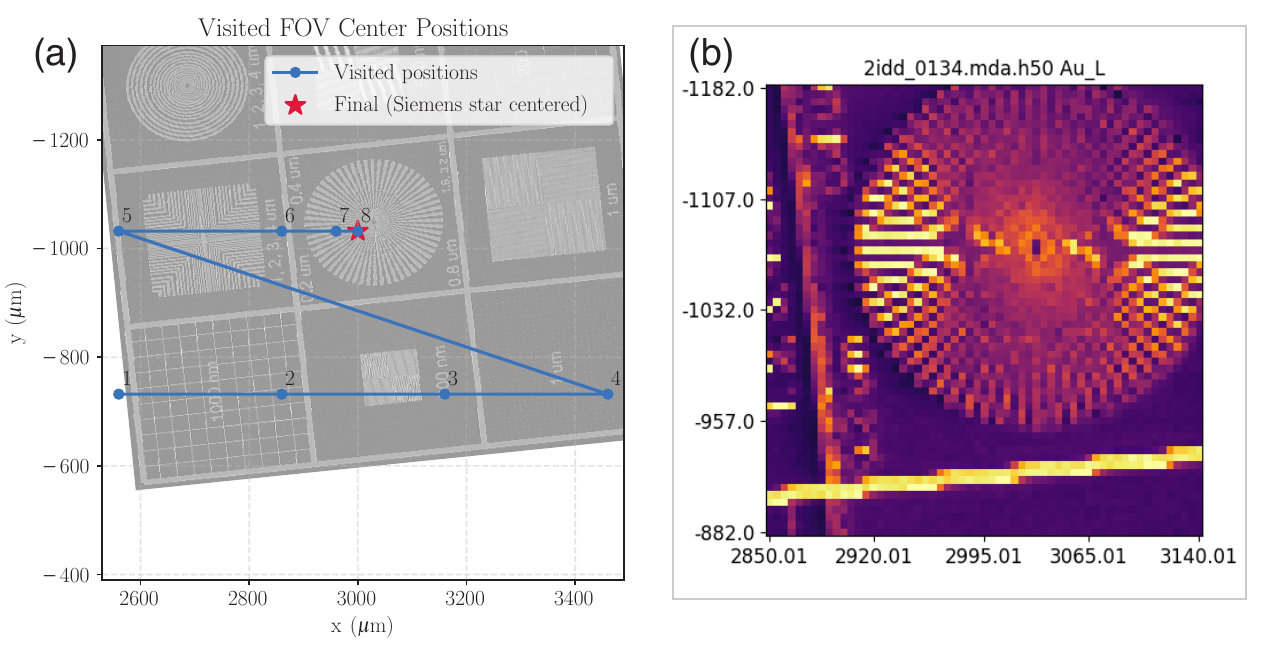}
    \caption{Trajectory and result of the feature search task. (a) The trajectory of the center of 2D image acquisitions in the sample space. Underlaid in the background is the SEM image of the sample (a test pattern), which is transformed to align with the trajectory plot. (b) The image returned by the last 2D scan tool call (the title and axis ticks are a part of the returned image), right before the agent concluded the process. The requested feature (Siemens star) is roughly centered in the FOV. }
    \label{fig:fov_search_result}
\end{figure}

We now demonstrate an automated feature search process conducted by EAA, where the agent was given tools to acquire images at given locations of the sample and was instructed to look for a feature described by natural language. 

The sample used for this experiment is a designed lithographic test pattern consisting of micron-scale gold features on a silicon nitride membrane. A scanning electron microscopy (SEM) image of the test pattern, rotated and scaled to align with the space of sample stage positions when the sample is mounted in the beamline, is shown in the background of Fig.~\ref{fig:fov_search_result}(a). Due to the small feature size, one can hardly locate a particular feature (such as a Siemens star) with a coarse scan of the entire sample, especially given that the overall shapes of the features are not unique. A local fine scan is required to distinguish a feature, and many such scans may be needed to locate the desired feature in a large sample. This necessitates an automated workflow that performs local fine scans, examines the image, and moves the FOV to the next location until the desired feature is found. 

We prompted the agent to look for the feature described as the following: ``\texttt{the center of a Siemens star, which is a disk formed by a lot of radial spokes (the spokes must be radial. A disk formed by concentric circles is not a Siemens star).}'' The search range, FOV size, and scan step size are also suggested to the agent. We explicitly instructed the agent to start with a grid search, but to close in on the feature with arbitrary direction and step size when a part of it appears in the FOV. The trajectory of the search is shown in Fig.~\ref{fig:fov_search_result}(a). We also prompted the agent to explain every tool call made, which was found to help stabilize the tool call performance in a long process. Starting from a corner of the sample, the agent sequentially moves the FOV and examines the image, and ``saw'' a part of the Siemens star at the 6th step. It then reduced the step size to make the Siemens star better centered. At the 8th step, the agent collected the image shown in Fig.~\ref{fig:fov_search_result}(b), and concluded the process. The image shows that the majority of the Siemens star is in the FOV and its center is close to the middle of the FOV, though not exactly. As one can see in the video recording available in the supplementary materials, the center of the Siemens star is already visible in the FOV at step 7, and a human operator would be able to directly read out the coordinates of the Siemens star's center from the axis ticks and bring it to the exact center of the FOV in one step. This reveals a gap in the current VLM ability to perform vision-based tasks that require some degree of quantitative analysis compared to humans; however, the automation of the process before the final stage already represents a significant reduction in human effort. 

\subsection{Case study 3: interactive data acquisition}

\begin{figure}
    \centering
    \includegraphics[width=1\linewidth]{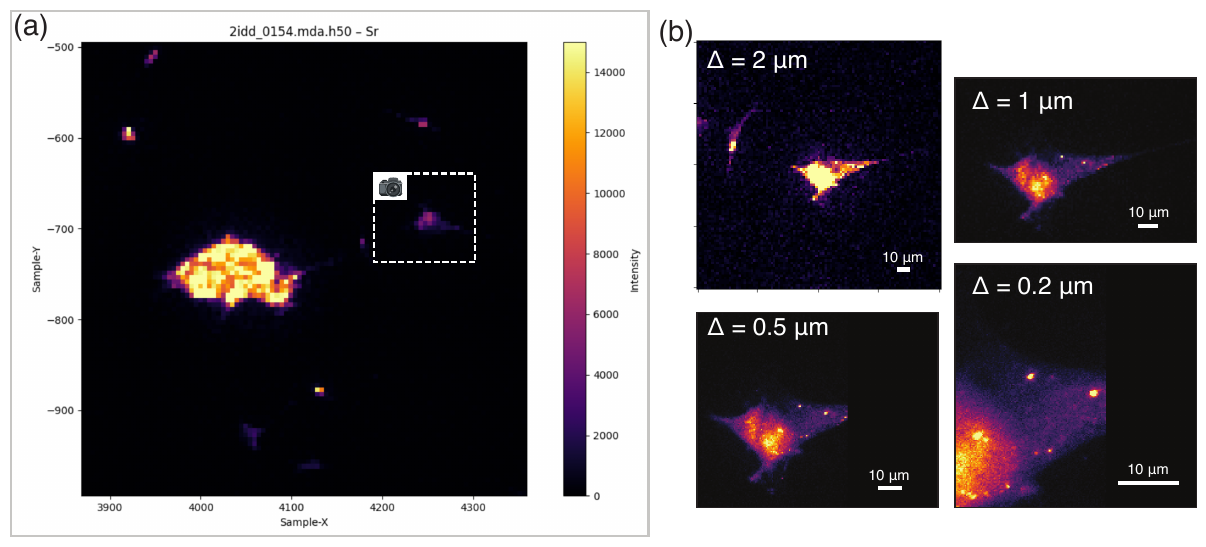}
    \caption{Input and acquired images during the interactive data acquisition session. (a) The image initially shown to the agent (agent did not see the dashed box). We subsequently took a screenshot of the area enclosed by the white dashed box and sent it to the agent, instructing it to take finer scans in that region. (b) Images captured by the agent with the 2D acquisition tool. These images are flipped vertically compared to (a) as (a) is a screenshot from the beamline control software and uses a different convention for plotting from the image acquisition tool of EAA. The scan step sizes are indicated at the top left corner. The image with $\Delta = 2$ \micron{} was acquired after the agent was instructed to perform a local scan of the user-selected feature. Images with smaller step sizes were acquired following the user's instruction to perform 3 scans sequentially. The truncation on the right of the images with $\Delta$ = 0.5 and 0.2 \micron{} is due to a problem in the hardware buffer that the tool depends on, irrelevant to EAA and the tools themselves.}
    \label{fig:interactive}
\end{figure}

While the examples above feature high-level automation of beamline operations with the objective of minimizing human involvement, EAA can also interact with human users, receiving and executing tasks dynamically. This capability provides a friendly chat interface for users to conduct measurements at complex instruments, handles simple calculations and planning prior to data acquisition, and also offers some degree of automation. We demonstrate this capability with the XRF setup at beamline 2-ID-D. 

The sample measured in this experiment is a CLN canine lymph node cell specimen treated with \ce{SrCl2}, mounted on a silicon nitride membrane. In EAA's WebUI, we sent the image shown in Fig.~\ref{fig:interactive}(a) (not including the dashed box) to the agent, followed by a screenshot in the white dashed box conveniently pasted into the chat box. We instructed the agent to take a local scan around the feature in the screenshot. No coordinates were mentioned in the text message. To determine the positions and size for the scan, the agent needed to associate the screenshot with the image of the full FOV, and figure out the location and size for the scan using the axis ticks. Despite the challenge, the agent made the correct 2D scan tool call at $x$ = 4250 and $y$ = -700, exactly at the center of the feature. The acquired image is shown in Fig.~\ref{fig:interactive}(b) (scan step size $\Delta$ = 2 \micron{}), where the target feature is clearly present and well centered. The image is vertically flipped compared to (a) because the latter is a screenshot from the beamline control software which uses a different plotting convention from the tools of EAA. Subsequently, we instructed the agent to tighten the FOV around the feature, and conduct three scans in a row with step size $\Delta$ = 1, 0.5, and 0.2 \micron{}. Specifically, we required that the scan should not collect more than 200 points in either $x$ or $y$. The agent determined that to achieve a tighter bound, the width of the FOV should be reduced to 120 \micron{} and the height should be reduced to 80 \micron{}. It then calculated the number of points in both directions, noting that the requirement on the number of points is satisfied for $\Delta$ = 1 \micron{}. For $\Delta$ = 0.5 and 0.2 \micron{}, the number of points would exceed 200 with the same FOV size, so the agent reduced the scan range to cap the number of points at 200 in both directions. The agent then launched 3 tool calls sequentially without human intervention, which collected the rest three images in Fig.~\ref{fig:interactive}(b). This experiment reveals the adequate capability of VLMs in experiment planning and text-based reasoning. 

\section{Discussion}
\subsection{Choice of VLM}

EAA is designed to work with any VLM as long as an OpenAI-compatible API is provided by the model's inference endpoint. Since OpenAI-compatible API format is widely accepted in the industry, the open-source community, as well as self-hosted LLM deployment tools (\emph{e.g.}, LMStudio and Ollama), this requirement is often readily satisfied. EAA also has a built-in adapter for the API of Ask Sage (asksage.ai) and Argo (argo.anl.gov); Both are LLM inference platform providing access to models from multiple vendors including OpenAI, Anthropic, and Google. Argo is developed at Argonne National Laboratory. Given the many choices, the decision of the most appropriate model to use should be made based on the complexity of the task, the requirement of response speed, the costs, and data privacy concerns (for commercial and third-party hosted models). We conducted a simple experiment assessing the task-performing accuracy and response latency of a few commonly available VLMs. However, we must note that with the rapid evolution of LLMs, the results of this study are expected to quickly be made obsolete by new models with much stronger capabilities, and the main purpose of the presented study is only to illustrate by how much models' costs and performance can differ. 

In the first task, we created a dummy image acquisition tool, and issued the following prompt to the model:

\texttt{Using the tool given to you, please acquire 4 images at 4 different locations in the sample. The 4 locations are arranged in a square grid. The top left image is at x = 0, y = 0, and the images are separated by 100 pixels in x or y. Each image should have a size of (256, 256) pixels. **IMPORTANT**: When making tool calls, make only one call at a time. Do not make multiple calls at once. When you finish acquiring the 4 images, say "TERMINATE".}

\begin{figure}
    \centering
    \includegraphics[width=1\linewidth]{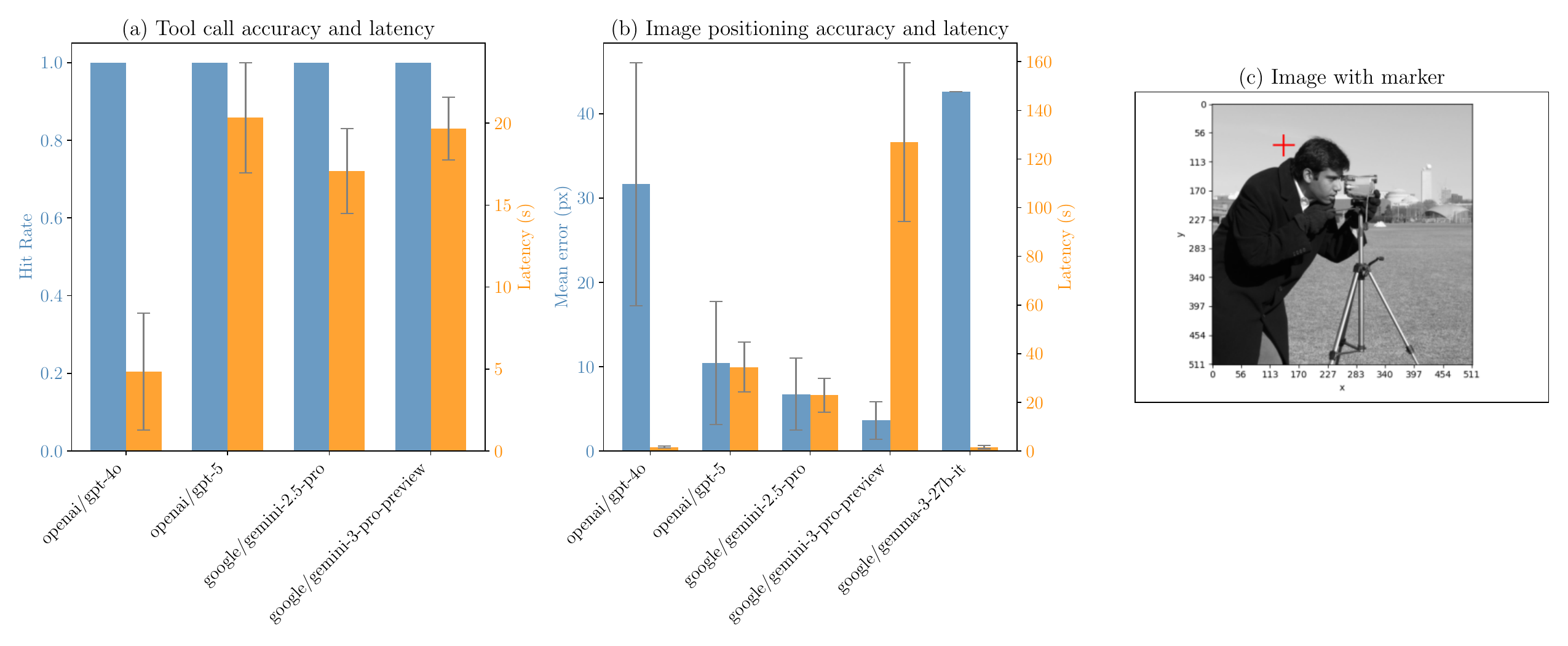}
    \caption{Model benchmarking results for two tasks related to experiment automation and image analysis. The bar heights indicate the average, and the half lengths of the error bars indicate the standard deviations of the measured quantities calculated over 10 trials for each model. Temperature and reasoning effort for model prediction was set to 1.0. (a) The performance of GPT-4o, GPT-5, Gemini 2.5 Pro, and Gemini 3 Pro Preview when prompted to acquired 4 images on a square grid using a provided image acquisition tool. ``Hit rate'' is the proportion that the $4\times 10$ expected tool calls are made. ``Latency'' shows the average time that the agent took to finish the task in each trial. (b) The accuracy of identifying the position of a marker in an image, tested on GPT-4o, GPT-5, Gemini 2.5 Pro, Gemini 3 Pro Preview, and Gemma 3 (27B). The accuracy is measured by the error of the reported coordinates compared to the true coordinates. (c) shows the image with the marker that AI sees.}
    \label{fig:model_benchmark}
\end{figure}

The test was performed on 4 tool calling-capable VLMs, namely GPT-4o, GPT-5, Gemini 2.5 Pro, Gemini 3 Pro Preview. GPT-5 and Gemini models were tested with their reasoning capability enabled. The temperature for inference was 1.0, and the default reasoning effort was used. For each model, 10 trials were conducted, and the statistics of the results are shown in Fig.~\ref{fig:model_benchmark}(a). The heights of the bars show the average of the measured quantities over the 10 trials, and the half lengths of the error bars indicate the standard deviation. Since the model is expected to make 4 calls to acquire the 4 images in each trial, there are 40 expected calls over the 10 trials. ``Hit rate'' in the figure shows the proportion that the 40 expected tool calls are made. ``Latency'' is the total time taken by the LLM to respond in a task, averaged over all trials. The plot shows that all the models attained 100\% hit rate, meaning they performed the 4 image acquisitions correctly over the 10 trials. However, GPT-5 and Gemini models had significantly larger latencies compared to GPT-4o because their reasoning processes took longer. This experiment indicates that when a model is capable enough for a task, a non-reasoning model can provide comparable accuracy compared to reasoning ones with faster speed. 

In the second task, the models were presented with an image with a red crosshair marker [Fig.~\ref{fig:model_benchmark}(c)]. The image has axis ticks from which the position of the marker can be read. The following prompt was sent to the models:

\texttt{The image in this message has a reticle marker. Read out the coordinates of the center of the reticle using the axis ticks. Report your answer in the format of x = \textlangle x\_coord\textrangle, y = \textlangle y\_coord\textrangle{} (for example, x = 20, y = 30). Only respond with the coordinates, no other text.}

The test was performed on GPT-4o, GPT-5, Gemini 2.5 Pro, Gemini 3 Pro Preview, and Gemma 3 (27B parameters). Since quantitative vision capability is needed for this task, vision reasoning becomes more important: the reasoning-capable VLMs, GPT-5 and Gemini, attained an average error around or below 10 pixels. Considering that the crosshairs of the marker are not aligned with axis ticks and the models were not given any tools to help with their analysis, this accuracy is considerably impressive. This, however, comes at the cost of latencies more than 10 times that of the non-reasoning models. Particular, Gemini 3 Pro Preview achieved a remarkable mean error of less than 5 pixels at the cost of a large latency of over 30 seconds due to the long reasoning process. GPT-4o attained an average position error of 32 pixels, and the open-weight, much smaller Gemma-3-27B had an average error of around 43 pixels. These models are faced with more challenges in tasks that require precision, but providing auxiliary tools will effectively compensate the shortcomings in the models' native capability. Meanwhile, open-weight models like Gemma 3 are locally deployable, which incurs minimal operation cost (when computing resources are already available) and prevents the risk in data privacy and intellectual property protection. 

\subsection{Safety guardrails}

To prevent instrument damage and safety hazards due to erroneous tool calls, we enforce guardrails at multiple levels. First, in tool functions that interact with experiment instruments, hard checks are implemented to ensure the input parameters such as motor positions are in reasonable ranges. Exceptions are thrown when the limits are violated, which are then picked up by the agent for it to revise its tool calls accordingly. Second, for all the tools, users can optionally require tool calls to be approved by a human before they are executed. Python and Bash coding tools which carry a higher risk have this requirement in place by default. Third, to prevent unwanted access or modification of the file system of the computer where EAA runs, EAA allows tools capable of interacting with the file system (examples are again Python and Bash tools) to execute in a container. The containerization strategy can also be used to restrict the network access of certain tools.

\section{Methods}

This section provides implementation-level details that complement the high-level system description given earlier.

\subsection{Agent implementation}

The agent class encapsulates a VLM client for managing connections to inference endpoints and a tool manager that maintains a registry of callable tools together with their schemas. Tool schemas describe names, input and output types, and semantic intent in a format understood by VLM servers. These schemas are transmitted alongside messages, enabling the VLM to select and invoke tools. Tool calls returned by the VLM are parsed, dispatched to the appropriate tool instances, and executed with model-supplied arguments. Execution results are captured by the task manager, appended to the conversational context, and returned to the VLM.

To support multimodal outputs under OpenAI-compatible APIs that do not permit images in tool responses, image-producing tools persist their outputs to disk and return file paths. The task manager performs a rule-based post-processing step that loads the images, encodes them as \emph{base64} for HTTP transport, and injects them into the context as auxiliary user messages immediately following the corresponding tool message. This preserves multimodal continuity while remaining compliant with API constraints.

Optional long-term memory is implemented via retrieval-augmented generation (RAG). Notable user messages are detected, embedded into a latent space, and stored in a vector database. During subsequent interactions, the top-$k$ most similar records are retrieved using cosine similarity and appended to the context, enabling cross-session recall of frequently accessed operational knowledge.

\subsection{Tool design and MCP integration}

EAA tools are implemented as Python classes to support persistent internal state, such as acquisition counters or parameter histories, across invocations. Within EAA, tools are typically invoked via function calling for low-latency, in-memory execution. For interoperability, any tool class can be wrapped as an MCP server, allowing it to be consumed by external MCP clients. Conversely, EAA can ingest external MCP servers through a generic \texttt{MCPTool} interface, exposing them to the agent as native tools.

This bidirectional compatibility allows EAA to participate in a broader agent–tool ecosystem while preserving the option of stateful, in-process execution when required for instrument control robustness and reliability.

\subsection{Sample preparation}

The lithographic test pattern used in the feature search experiment was fabricated by patterning a 1-\micron{}-thick gold layer on a silicon nitride membrane with a smallest feature size of 200 nm. The lithography layout was generated using the CNST Nanolithography toolbox. Fabrication was carried out at the Center for Nanoscale Materials (CNM) at Argonne National Laboratory using an electron-beam lithography and electroplating process. Poly(methyl methacrylate) (PMMA) resist was spin-coated onto the substrate and patterned with a JEOL JBX-8100 FS electron-beam lithography system. The exposed resist was developed to form a mold, which was subsequently filled by gold electroplating. After plating, the resist was removed and the sample was prepared for X-ray microscopy measurements.

The biological sample used in the interactive data acquisition experiment was a CLN canine lymph node cell specimen treated with \ce{SrCl2}. The cells were seeded onto silicon nitride membranes (1.5 $\times$ 1.5~mm$^2$). A 50~$\mu$L droplet of cells suspended in growth medium was deposited onto the membrane, followed by treatment with 300~$\mu$L of fresh medium containing 40~mM \ce{SrCl2} prior to X-ray fluorescence microscopy measurements.

\section{Conclusion}

We designed EAA with the core principles and features of vision-capability, flexible workflows with custom levels of LLM involvement, and a modern, adaptable library and protocol of instrument-controlling tools. We demonstrated EAA in real-world tasks including automated microscope focusing, feature search, and an interactive session where the user issues provisional instructions. Not bound to a certain LLM or a specific pattern or workflow of operation, EAA is designed to be a future-proofing system that will keep up with the rapid evolution of artificial intelligence. We expect EAA to improve beamline efficiency, free up beamline scientists' time for creative works, and make beamline operations more friendly for users. These are also the objectives that we will constantly push for in the future.

\section*{Code availability}

The EAA project is located at \url{https://github.com/AdvancedPhotonSource/EAA}.

\section*{Acknowledgments}

We thank Matthew T.~Dearing, Vincent Chirio, and others in the Argo team for developing and supporting Argo, an LLM inference platform, which we used for the development and demonstration of EAA.

This research was performed at the Advanced Photon Source on APS beam time award (DOI: 10.46936/APS-191512/60015223) and the Center for Nanoscale Materials, both U.S.~Department of Energy (DOE) Office of Science user facilities, and is based on work supported by Laboratory Directed Research and Development (LDRD) funding from Argonne National Laboratory (grant No.~2025-0496), provided by the Director, Office of Science, of the U.S.~DOE under Contract No.~DE-AC02-06CH11357. JP acknowledges support from the National Institutes of Health NIH/NIAID, award 1P01AI165380-01; Multi-Scale Evaluation and Mitigation of Toxicities Following Internal Radionuclide Contamination. MJC also acknowledges support from the U.S. Department of Energy, Office of Science, Office of Advanced Scientific Computing Research and Office of Basic Energy Sciences, Scientific Discovery through Advanced Computing (SciDAC) program under the MIRAGE project.

\printbibliography

\end{document}


\textbf{GOVERNMENT LICENSE}

The submitted manuscript has been created by UChicago Argonne, LLC, Operator of Argonne
National Laboratory (``Argonne''). Argonne, a U.S. Department of Energy Office of Science laboratory, is operated under Contract No. DE-AC02-06CH11357. The U.S. Government retains for
itself, and others acting on its behalf, a paid-up nonexclusive, irrevocable worldwide license in
said article to reproduce, prepare derivative works, distribute copies to the public, and perform
publicly and display publicly, by or on behalf of the Government. The Department of Energy will
provide public access to these results of federally sponsored research in accordance with the DOE
Public Access Plan. http://energy.gov/downloads/doe-public-access-plan.

\maketitle

\section{Video recordings of case studies}

\subsection{Case study 1: microscope focusing}

\url{https://anl.box.com/s/jqz45hgfcvqkx0ji8b7ma9fvqd9ozkyb}

\subsection{Case study 2: searching for desired features}

\url{https://anl.box.com/s/n1ws29mzld9pxkzmz0kc4x64r5h32s49}

\subsection{Case study 3: interactive data acquisition}

\url{https://anl.box.com/s/b2dc6cormgyqrr6j462kg2kn4hyrqlzk}